\documentclass[conference]{IEEEtran}
\IEEEoverridecommandlockouts
\usepackage{cite}
\usepackage{amsmath,amssymb,amsfonts}
\usepackage{algorithmic}
\usepackage{graphicx}
\usepackage{textcomp}
\usepackage{colortbl}
\usepackage{xcolor}
\usepackage[hidelinks]{hyperref}
\usepackage{caption}
\usepackage{subcaption}
\usepackage{enumitem}
\usepackage{url}  % For handling URLs

\def\BibTeX{{\rm B\kern-.05em{\sc i\kern-.025em b}\kern-.08em
    T\kern-.1667em\lower.7ex\hbox{E}\kern-.125emX}}

\begin{document}

\title{Improving Long-Tailed Object Detection with Balanced Group Softmax and Metric Learning}

\author{\IEEEauthorblockN{Satyam Gaba}
\IEEEauthorblockA{\textit{Univeristy of the Cumberlands} \\
\textit{Williamsburg, US} \\
sgaba27353@ucumberlands.edu}
% \and
% \IEEEauthorblockN{2\textsuperscript{nd} David Ostrowski}
% \IEEEauthorblockA{\textit{School Of Professional Studies. Northwestern University} \\
% \textit{Evanston, IL}\\
% david.ostrowski@northwestern.edu}
% \and
}

\maketitle

\begin{abstract}
Object detection has been widely explored for class-balanced datasets such as COCO. However, real-world scenarios introduce the challenge of long-tailed distributions, where numerous categories contain only a few instances. This inherent class imbalance biases detection models towards the more frequent classes, degrading performance on rare categories. In this paper, we tackle the problem of long-tailed 2D object detection using the LVISv1 dataset, which consists of 1,203 categories and 164,000 images. We employ a two-stage Faster R-CNN architecture and propose enhancements to the Balanced Group Softmax (BAGS) framework to mitigate class imbalance. Our approach achieves a new state-of-the-art performance with a mean Average Precision (mAP) of 24.5\%, surpassing the previous benchmark of 24.0\%.

Additionally, we hypothesize that tail class features may form smaller, denser clusters within the feature space of head classes, making classification challenging for regression-based classifiers. To address this issue, we explore metric learning to produce feature embeddings that are both well-separated across classes and tightly clustered within each class. For inference, we utilize a k-Nearest Neighbors (k-NN) approach to improve classification performance, particularly for rare classes. Our results demonstrate the effectiveness of these methods in advancing long-tailed object detection.

\end{abstract}

%%%%%%%%% BODY TEXT
\section{Introduction}
Object detection is a fundamental yet challenging task in computer vision, with applications ranging from autonomous driving to surveillance and medical imaging. An object detection pipeline typically consists of the following key steps:

\begin{itemize} 
    \item Identifying regions of interest by generating bounding boxes
    \item Extracting features for each bounding box 
    \item Classifying objects and refining bounding box coordinates using the extracted features 
\end{itemize}

Faster R-CNN \cite{ren2016faster} revolutionized object detection by introducing the Region Proposal Network (RPN), which generates region proposals and shares full-image convolutional features with the detection head. This architecture laid the foundation for two-stage detectors, which process images in two phases: region proposal and classification with bounding box refinement. In contrast, single-stage detectors like SSD \cite{liu2016ssd} and YOLO \cite{redmon2016you} predict class probabilities and bounding box coordinates directly from a pre-defined set of anchor boxes in a single forward pass. While single-stage detectors are computationally faster, two-stage detectors generally offer higher accuracy, making them preferable for applications requiring precise detection

Current research in object detection has achieved significant progress, primarily on class-balanced datasets such as PASCAL VOC \cite{everingham2010pascal} and COCO \cite{lin2014microsoft}. However, real-world scenarios present a more complex challenge: object categories tend to follow a long-tailed distribution, where a small number of frequent categories dominate the dataset, while many categories contain only a few instances. This class imbalance introduces significant bias in detection models, favoring frequent classes at the expense of rare ones.

Despite the growing interest in improving object detection models, the problem of handling long-tailed distributions remains relatively under-explored. Developing efficient models that can maintain performance across both frequent and rare categories is essential for real-world applications, as imbalanced datasets are the norm rather than the exception. In this work, we focus on two-stage detectors and propose methods to address the challenges posed by long-tailed distributions, leveraging the LVIS dataset, which is specifically designed to reflect real-world class imbalance.

The challenge of long-tailed object detection remains under-explored, yet it is crucial for applications in the real world, where imbalanced datasets are the norm. In this work, we address the challenges posed by long-tailed distributions using LVISv1 \cite{gupta2019lvis}, a dataset specifically designed to capture this imbalance, with 1,203 categories and 164,000 images. We adopt the Faster R-CNN framework and explore novel enhancements to mitigate class imbalance. In particular, we improve upon Balanced Group Softmax (BAGS) \cite{li2020overcoming}, a state-of-the-art approach to long-tailed detection by tackling group based imbalances. Furthermore, we explore metric learning techniques to improve feature embeddings and ensure better classification for rare classes.

The rest of the paper is structured as follows: Section II reviews related work, Section III details our proposed methods, and Section IV presents the experimental results and analysis.

\section{Related Work}

The LVIS dataset \cite{gupta2019lvis} was introduced to address the challenges of long-tailed object detection and instance segmentation. Standard object detection models struggle with this dataset because frequent head classes dominate tail classes during training. Several recent studies have attempted to address this issue by modifying the loss function to reduce class imbalance. For example, Focal Loss \cite{lin2017focal} assigns higher weights to less confident predictions, while Equalization Loss \cite{tan2020equalization} reduces the weight suppression of tail classes by head classes. While these methods outperform naive models, the performance gap between balanced and long-tailed datasets persists.

Another important work by Kang et al. \cite{kang2019decoupling} proposed that class imbalance primarily affects classification rather than feature representation. Their method decouples feature learning from classifier training: first, the model learns feature representations using a standard model, and then the feature extraction layers are frozen while the classifier head is fine-tuned with normalized weights to balance head and tail classes. While effective, this strategy relies heavily on the assumption that tail categories share visual similarity with head categories, which may not always hold true.

In Learning to Segment \cite{learning2segment}, the authors propose transferring knowledge from head to tail classes by grouping categories based on frequency and training the model in phases. Each group is trained sequentially, with balanced replay of earlier groups to prevent forgetting. The classifier weights for tail classes are learned as a linear combination of head class weights. However, the performance of this method depends on the assumption that tail classes resemble head classes, which limits its applicability.

More recently, Li et al. \cite{li2020overcoming} introduced Balanced Group Softmax (BAGS), where categories are divided into bins based on frequency, and softmax loss is applied within each bin. This strategy ensures that rare classes compete only with others of similar frequency, alleviating some effects of class imbalance. Although BAGS achieves state-of-the-art performance, we identified areas for improvement. Specifically, the logic for bin selection was not thoroughly explored, and class imbalance within each bin remains an issue.

To build upon BAGS, we conduct extensive ablation studies to explore different binning strategies. We propose novel clustering-based methods for bin creation and experiment with weighted variants of BAGS that apply popular loss weighting techniques within each bin. Our improvements yield an average precision (AP) boost of 1.25\% for rare and common classes.

However, simply mitigating imbalance within bins is insufficient to fully solve the long-tailed detection problem. To further enhance detection, we explore metric learning, a technique that aims to learn feature embeddings with minimal intra-class distance and maximal inter-class separation. We integrate Center Loss \cite{wen2016discriminative} and Cosine Loss (CosFace) \cite{wang2018cosface} into our training to ensure tighter class clusters. Additionally, we introduce a novel Euclidean Cross-Entropy Loss, which directly minimizes intra-class feature distances. For inference, we experiment with both k-Nearest Neighbors (k-NN) and traditional cross-entropy classification to evaluate the effectiveness of our embeddings.

These enhancements result in improved performance, particularly for tail classes, demonstrating the potential of combining metric learning with enhanced loss functions for long-tailed object detection.

\section{Method}
\subsection{Baseline Model - Faster-RCNN}

The Faster R-CNN network architecture is composed of three primary components, each performing a distinct task: the backbone network for extracting feature maps from the input image, the Region Proposal Network (RPN) for generating region proposals, and the ROI head network, which is responsible for bounding box regression and object classification. \par

\begin{figure}[ht]
    \centering
    \includegraphics[width=0.3\textwidth]{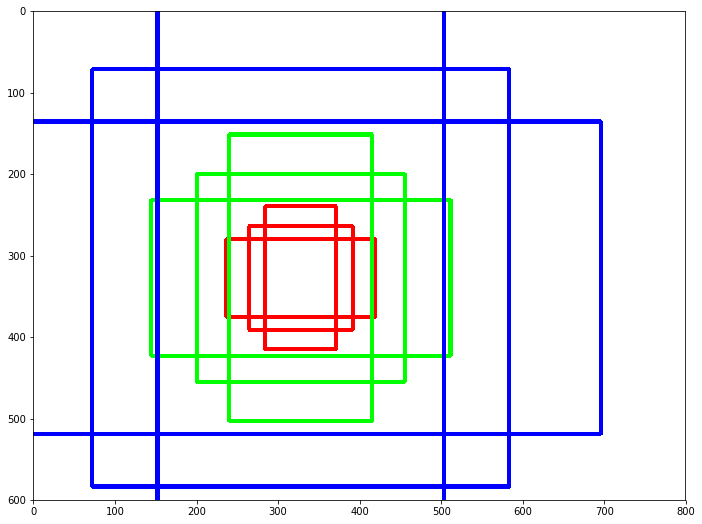}
    \caption{Anchor boxes of three different sizes and height to width ratio at an anchor point}
    \label{fig:anchors}
\end{figure}

The backbone network is typically a convolutional neural network, such as VGG16 or ResNet50, designed to extract feature maps from the input image. ResNet backbones are often preferred over VGG due to their ability to address the vanishing gradient problem in deeper networks. The backbone network generates a set of anchors uniformly distributed across the entire image. Initially, the network defines a grid of anchor points on the image, and for each anchor point, it generates anchor boxes with varying scales, sizes, and aspect ratios, as illustrated in Figure~\ref{fig:anchors}. \par

The features extracted by the backbone model, combined with the anchor boxes, are processed by the Region Proposal Network (RPN) to generate object proposals. The RPN outputs a set of bounding box proposals, which are subsequently evaluated by a classifier and a regressor in the Region of Interest (ROI) head to confirm object presence and refine the predictions. Specifically, the RPN determines whether an anchor box corresponds to the foreground or background and adjusts the anchor's coordinates.\par

The RPN generates a fixed number of proposals per image. The ROI head selects the top proposals with the highest probabilities of containing objects. It comprises two components: a classifier for object category prediction and a bounding box regressor for refining the coordinates of the bounding boxes. In the classifier head, the extracted features are passed through fully connected layers corresponding to the number of categories, and the Cross-Entropy loss is applied during training. In the bounding box regression head, anchor coordinates are adjusted to better match the ground truth using the Smooth L1 loss. \par

Faster R-CNN can be trained using two methods: (1) alternating training of the RPN and the final classifier and regressor, or (2) joint training of all components simultaneously. This work adopts the latter approach, which is computationally more efficient while maintaining comparable accuracy. In both approaches, gradients are propagated back to the convolutional neural network (CNN) to update shared weights. \par

\subsection{Decoupling Classifier}

It has been observed that the bias of a classifier towards head classes arises from imbalances in the weight norms of the classifier. Specifically, the weight norms corresponding to head classes are significantly higher than those of tail classes \cite{kang2019decoupling}. Decoupling classifier approaches posit that this imbalance primarily affects classification, while feature representation learning remains unaffected. These approaches typically involve training the feature extraction network with a standard cross-entropy loss, freezing the feature extraction layers, and re-training only the classifier head to address the disparity in weight norms between head and tail classes. The re-training is conducted using a re-sampled, balanced data distribution.

Kang et al. \cite{kang2019decoupling} explore several strategies for sampling a balanced dataset for classifier re-training, including: (1) Instance-balanced sampling; (2) Class-balanced sampling; (3) Square-root sampling; and (4) Progressively-balanced sampling. 

Rebalancing methods for classifier weight norms, such as (1) Classifier Re-Training (cRT); (2) Nearest Class Mean classifier (NCM); (3) $\tau$-normalized classifier ($\tau$-normalization); and (4) Learnable Weight Scaling (LWS), have shown improved performance, particularly for tail classes. In our experiments, we selected $\tau$-normalization due to its superior performance and absence of training overhead \cite{kang2019decoupling}.

\subsubsection*{Tau normalization}
% Tau normalization does not require retraining the classifier. It uses the joint-trained model and reweighs at classifier weights during inference. This method normalizes the weight across all categories by dividing weight by their respective norms. There is a temperature value ($\tau$) between 0 to 1, that controls the amount of normalization. The updated classifier weights will be:
% \begin{equation*}
%     \hat{w} = w/|w|^{\tau}
% \end{equation*}
% Higher the value of $\tau$, more uniform distribution weights as shown in Figure \ref{fig:tau_norms}.\\
% With Tau-normalization there is no overhead during training or inference time.
% \begin{figure}
%     \centering
%     \includegraphics{figures/tau_norms.png}
%     \caption{Weight norms for each category with different values of $\tau$. Categories are sorted according to instance count}
%     \label{fig:tau_norms}
% \end{figure}

\begin{figure}[ht]
    \centering
    \includegraphics{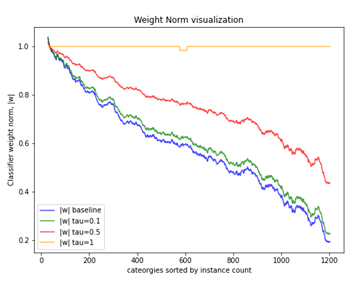}
    \caption{Weight norms for each category with different values of $\tau$. Categories are sorted by instance count.}
    \label{fig:tau_norms}
\end{figure}

$\tau$-normalization eliminates the need for re-training the classifier. Instead, it operates on the jointly-trained model and reweighs the classifier weights during inference. This method normalizes the weights across all categories by dividing each weight by its respective norm. A temperature parameter ($\tau$), ranging from 0 to 1, controls the extent of normalization. The updated classifier weights are computed as follows:
\begin{equation*}
    \hat{w} = \frac{w}{|w|^{\tau}}
\end{equation*}
A higher value of $\tau$ results in a more uniform distribution of weights, as illustrated in Figure \ref{fig:tau_norms}. 

With $\tau$-normalization, there is no computational overhead during either training or inference.

\subsection{Benchmark Model - Balanced Group Softmax}

The standard softmax loss, employed by the baseline Faster R-CNN model, is unsuitable for long-tailed detection. This is because the learning process for tail classes is overwhelmed by negative gradients generated by head classes, resulting in highly skewed classifier weight norms (which correlate with instance count) and poor performance for rare categories, often yielding a mean Average Precision (mAP) close to zero.

The Balanced Group Softmax (BAGS) loss \cite{li2020overcoming} addresses this issue by dividing categories into groups based on their frequencies and performing group-wise training. By grouping categories with similar frequencies, BAGS eliminates the dominance of negative gradients from head classes over tail classes.

Formally, the $C$ categories are divided into $N$ groups. A category $j$ is assigned to group $G_n$ if the instance count of category $j$ falls within the range $[s_n^l, s_{n+1}^l)$, where $s_n^l$ represents the lower bound on instance frequencies for group $G_n$. According to \cite{li2020overcoming}, the best results on the LVIS dataset were achieved with $N = 4$ and the following bounds:
\[
s_1^l = 0, \quad s_2^l = 10, \quad s_3^l = 100, \quad s_4^l = 1000.
\]
The background category is assigned to group $G_0$. Additionally, each group includes an "Others" category, which represents all categories not included in that group.

During training, for a proposal with ground truth label $c$, only two groups are activated: $G_0$ and $G_n$, where $c \in G_n$. Within these groups, the softmax cross-entropy loss is computed. To prevent imbalance within each batch, the number of "Others" samples is limited.

The network's output is represented as \( z \in \mathbb{R}^{(C+1)+(N+1)} \). The probability of category $j$ is computed as:
\[
p_j = \frac{e^{z_j}}{\sum_{i \in G_n} e^{z_i}}
\]
where $n$ is such that $j \in G_n$. In groups where category $c$ is not included, the "Others" class serves as the ground truth class.

The final loss is defined as:
\[
L_k = -\sum_{n=0}^N \sum_{i \in G_n} y_i^n \log(p_i^n)
\]
where $y^n$ and $p^n$ represent the label and predicted probability within group $G_n$.

\subsection{Extensions to BAGS Loss}\label{sec:bags_ext}

\begin{figure}[h!]
  \centering
  \includegraphics[width=\columnwidth]{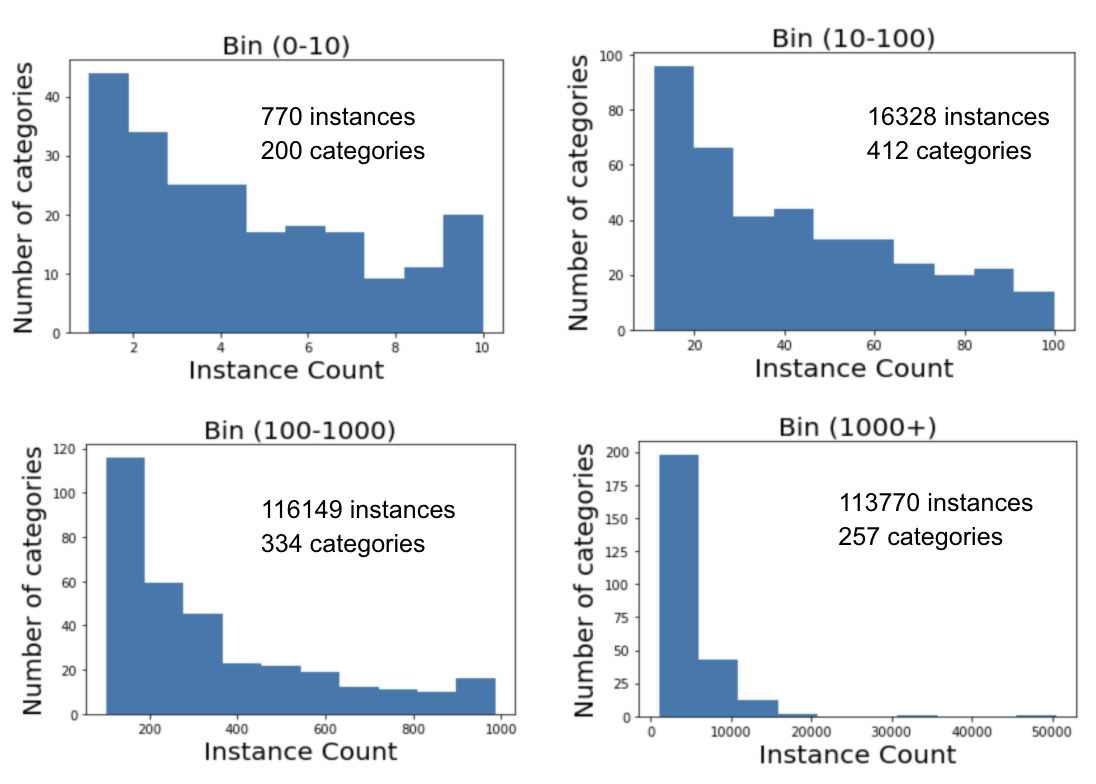}
  \caption{Class distribution imbalance within groups/bins generated by the Balanced Group Softmax method.}
  \label{fig:intra-bin-imbalance}
\end{figure}

The use of softmax within each bin independently, as implemented in BAGS, addresses inter-group class imbalance by preventing frequent classes from dominating infrequent ones during training. However, class imbalance persists within individual bins, as the dataset is large and the instance frequencies within a bin can vary significantly. This intra-bin imbalance is illustrated in Figure \ref{fig:intra-bin-imbalance}.

To address the intra-bin imbalance, we experimented with several methods, which are described in the following sub-sections.

\subsubsection{Altering the Number of Bins for BAGS}

In the original BAGS implementation, the optimal configuration consists of 4 bins, determined by thresholds of 10, 100, 1000, and 10,000 instance counts \cite{li2020overcoming}. A straightforward approach to mitigate intra-bin imbalance is to increase the number of bins. 

One experiment extended the model to 8 bins, but this led to issues of insufficient samples in some bins, which degraded performance. To avoid this problem, we adopted a more targeted strategy by splitting the bin with the highest number of samples—specifically, the bin with instance counts between 100 and 1000. This bin was divided into two smaller bins, corresponding to categories with instance counts of 100–500 and 500–1000.

After redefining the bins, we updated the mappings of labels to their respective bins and trained the classification head using the newly defined bin structure.

BAGS benefits from the group-wise training of categories by effectively isolating the training of tail classes from the influence of head classes. To further enhance classification performance, we experimented with different distributions of categories within bins as well as varying the number of bins. The objective of these experiments was to maximize the effectiveness of the separation achieved through group-wise training.

\subsubsection{Clustered BAGS}

In the original BAGS implementation \cite{li2020overcoming}, no ablation study was conducted to explore varying the frequency ranges for each bin. Decision boundaries such as 10, 100, and 1000 were selected without rigorous justification. We observed that the tail categories within each bin exhibited low Average Precision (AP) scores, suggesting that the long-tail distribution within bins negatively affected their performance. To address this, we introduced "Clustered BAGS," where image categories were clustered based on their instance frequencies. Additionally, we ensured that the clusters for tail frequencies contained fewer categories, reducing the adverse effects of long-tail distributions. The thresholds for the bins in our implementation were as follows: 0–22, 23–90, 91–1000, 1001–18,050, and 18,051+.

\subsubsection{BAGS with Class Weights}

To manage intra-bin imbalance, we applied a classic approach: class weighting. Each class was assigned a weight inversely proportional to its instance frequency, thereby reducing the impact of frequent categories within a bin. Formally, the initial weight for each class \(i\) was calculated as:
\[
w_{\text{init}}(i) = \frac{1}{\text{instance\_count}(i)}.
\]
These weights were normalized within each bin to ensure they summed to 1:
\[
w_{\text{normalized}}(i) = \frac{w_{\text{init}}(i)}{\sum_{j \in G_n} w_{\text{init}}(j)}, \quad i \in G_n.
\]
When no weights are used, the effective weight for each category is 1, and the total weight of categories in a group equals the number of categories in that group. To preserve the overall contribution of the loss term, the normalized weights were re-scaled by the number of categories in the bin:
\[
w_{\text{final}}(i) = w_{\text{normalized}}(i) \times |G_n|, \quad i \in G_n.
\]
These final weights were then applied during training.

\subsubsection{BAGS with Focal Loss}

Another approach to address intra-bin imbalance was to replace the standard cross-entropy loss with Focal Loss \cite{lin2017focal}. Focal Loss dynamically focuses more on harder-to-classify examples while reducing the importance of well-classified or easy examples. This inherently assigns greater weight to less frequent categories, as they are typically harder to classify. For head classes, this loss also prioritizes difficult-to-classify objects within those categories.

Focal Loss for binary classification is defined as:
\[
FL(p_t) = -(1 - p_t)^\gamma \log(p_t),
\]
where \(p_t\) is the predicted probability for the true class, \(\gamma\) is a tunable focusing parameter, and \(-\log(p_t)\) corresponds to the cross-entropy loss. Specifically:
\[
p_t = \begin{cases} 
p, & \text{if the label is 1}, \\
1 - p, & \text{if the label is 0}.
\end{cases}
\]
When a sample is misclassified (\(p_t\) is small), the modulating factor \((1 - p_t)^\gamma\) remains close to 1, assigning greater importance to the loss. Conversely, as \(p_t \to 1\), the factor approaches 0, effectively down-weighting the loss for well-classified examples.

\subsection{Metric Learning}

\begin{figure}[h!]
  \centering
  \includegraphics[width=0.7\columnwidth]{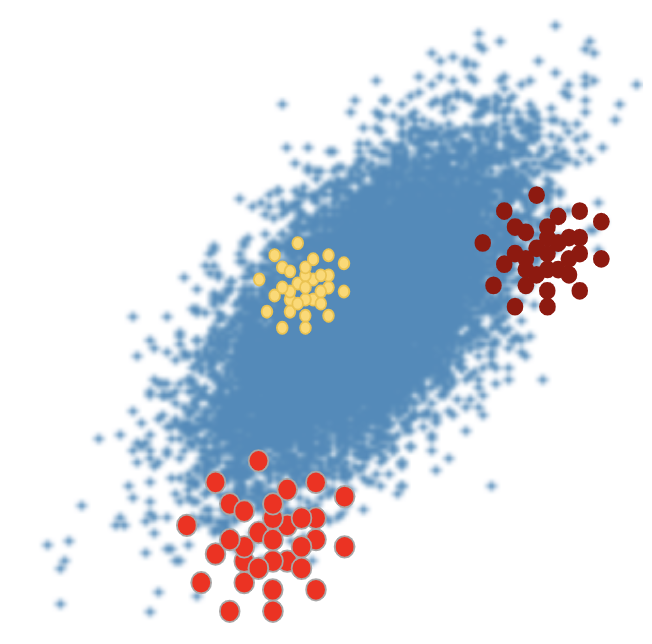}
  \caption{Tail class feature clusters occurring within or at the boundary of head class feature clusters. Blue dots represent the head class, while other colors represent tail classes.}
  \label{fig:metric-learning-hypothesis}
\end{figure}

Object detection involves both classification and bounding box regression. However, features of tail classes may form smaller clusters near or within the larger feature clusters of head classes, as depicted in Figure \ref{fig:metric-learning-hypothesis}. The final layer of the classification head learns a linear decision boundary based on these features, which might be insufficient for effective classification. In such cases, nearest neighbor classification could be more suitable. For nearest neighbor classification to work effectively, features of different categories must be well-separated. To achieve this, we explore Center Loss \cite{wen2016discriminative} and Large Margin Cosine Loss \cite{wang2018cosface}, described in the following subsections.

\subsubsection{Center Loss}

Softmax cross-entropy loss aims to separate features of different classes but does not ensure that feature clusters are tight. Center Loss addresses this by maintaining a feature center for each class and minimizing the distance between features and their corresponding class centers. Formally, Center Loss is defined as:
\[
L_{\text{Center}} = \frac{1}{2}\sum_{i=1}^{m}\|x_i - c_{y_i}\|_2^2,
\]
where \(c_{y_i}\) represents the center of the \(y_i\)-th class's features. The class centers are updated incrementally during training using the mini-batch rather than computing them over the entire dataset \cite{wen2016discriminative}.

The model is trained with a combination of softmax cross-entropy and center loss:
\[
L = L_{\text{softmax}} + \lambda L_{\text{Center}}.
\]
We perform inference using two methods:
\begin{itemize}
    \item \textbf{Method 1:} Arg-max over softmax scores (ignoring the centers).
    \item \textbf{Method 2:} K-Nearest Neighbor (KNN) inference, where we compute class centers by maintaining running means of features. For a given bounding box, the score for a class is the negative distance from the corresponding center. Softmax is applied to these scores to compute class probabilities.
\end{itemize}

\subsubsection{Large Margin Cosine Loss}

Losses like Center Loss outperform conventional softmax loss in tasks such as facial recognition by improving feature discrimination. These losses aim to maximize inter-class variance while minimizing intra-class variance. Large Margin Cosine Loss (LMCL) \cite{wang2018cosface} reformulates the softmax loss as a cosine loss by \(L_2\)-normalizing both feature and weight vectors to remove radial variations and introducing a cosine margin term to maximize the decision margin in angular space.

LMCL is defined as:
\[
L_{\text{LMCL}} = \frac{1}{N} \sum_{i} -\log \frac{e^{s(\cos(\theta_{y_i, i}) - m)}}{e^{s(\cos(\theta_{y_i, i}) - m)} + \sum_{j \neq y_i} e^{s \cos(\theta_{j, i})}},
\]
where \(s\) is a scale factor, \(\theta_{y_i, i}\) is the angle between the feature vector and the weight vector for the true class, and \(m\) is the margin term. 

\begin{figure}[h!]
    \centering
    \includegraphics[width=0.5\textwidth]{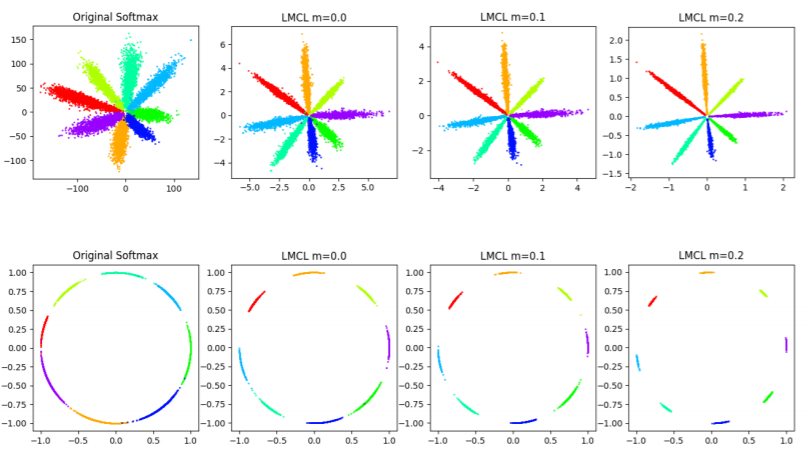}
    \caption{A toy experiment comparing different loss functions on 8 identities with 2D features. The first row maps features to Euclidean space, and the second projects them onto angular space. The gap widens as the margin \(m\) increases \cite{wang2018cosface}.}
    \label{fig:cos_loss_vis}
\end{figure}

LMCL reduces intra-class variance and increases inter-class variance by normalization and angular margin maximization, as illustrated in Figure \ref{fig:cos_loss_vis}.

\subsubsection{Euclidean Cross Entropy Loss}

Euclidean Cross Entropy (ECE) Loss modifies the numerator in the softmax term of Cross Entropy Loss, which traditionally uses the unnormalized cosine distance (\(w_i x_i\)) between weight and feature vectors. Instead, ECE Loss uses the Euclidean distance between learned parameters \(w_{y_i}\) and feature vectors. It is defined as:
\[
L_{\text{ECE}} = - \sum_{i=1}^{m} \log \left( \frac{e^{\frac{-\|x_i - w_{y_i}\|^2}{t}}}{\sum_{j=1}^{m} e^{\frac{-\|x_i - w_{y_j}\|^2}{t}}} \right),
\]
where \(t\) is a temperature scaling parameter. By replacing cosine similarity with Euclidean distance, ECE Loss provides a different approach to measure the relationship between weights and features.

\section{Experiments}

\subsection{Reproduced Baseline and SOTA}
We trained the Faster R-CNN model on the LVIS dataset for 12 epochs using 4 GPUs, each with a batch size of 2 images. The learning rate was set to 0.025, as recommended by the MMDetection library. The overall mean Average Precision (mAP), as well as mAP for rare, common, and frequent categories, were evaluated.

Furthermore, we trained the Balanced Group Softmax (BAGS) model on the LVIS-V1 dataset. The results are presented in Table \ref{table:reproduce}. The BAGS model demonstrated an improvement in overall mAP and nearly doubled the mAP for rare categories. While the original BAGS paper focused on LVIS-V0.5, our results on LVIS-V1 remain comparably robust.

\begin{table}[!h]
\centering
\begin{tabular}{|c|c|c|c|c|}
\hline
          & Overall & Rare & Common & Frequent \\ \hline
Baseline  & 21.1    & 8.1  & 19.6   & 28.6     \\ \hline
BAGS      & 24.0    & 15.6 & 22.8   & 29.2     \\ \hline
\end{tabular}
\caption{mAP results on the LVIS-V1 validation set for the baseline Faster R-CNN and the BAGS model.}
\label{table:reproduce}
\end{table}

\subsection{Decoupling - Tau Normalization}
We analyzed the impact of varying $\tau$ values during inference, applying $\tau$-normalization to the baseline model to enhance performance for tail classes. Table \ref{table:tau_norm} illustrates the overall mAP and AP for rare categories (APr). Increasing $\tau$ led to an improvement in APr, albeit at the cost of reduced overall mAP. This trade-off arises because enhancing performance on tail categories diminishes performance for head categories, which dominate the dataset and have a greater influence on overall mAP.

\begin{table}[!h]
\centering
\begin{tabular}{|l|l|l|}
\hline
$\tau$       & mAP                           & APr                          \\ \hline
0 (baseline) & \cellcolor[HTML]{9E9E9E}21.1 & 0.9                          \\ \hline
0.05         & 20.0                          & 0.9                          \\ \hline
0.1          & 19.8                          & 1.1                          \\ \hline
1            & 12.8                          & \cellcolor[HTML]{9E9E9E}4.0  \\ \hline
\end{tabular}
\caption{Effect of $\tau$-normalization on the validation set. While mAP decreases with higher $\tau$, AP for rare categories (APr) improves.}
\label{table:tau_norm}
\end{table}

\subsection{Ablation Studies: Random BAGS}
To assess the impact of sorting categories by instance frequency prior to binning, we performed an ablation study where the bin sizes and the total number of bins were fixed, but the category distribution was randomized.

Table \ref{table:rags} compares the performance of the Random BAGS (RAGS) model with the original BAGS model. The AP for rare categories dropped significantly in the RAGS model, as tail categories were suppressed regardless of their assigned bins. Conversely, frequent category AP increased due to their distribution across bins, allowing them to dominate other categories without intra-group competition.

\begin{table}[!h]
\centering
\begin{tabular}{|l|l|l|l|l|}
\hline
\textbf{}     & Overall & Rare & Common & Frequent                     \\ \hline
BAGS          & 24.0    & 15.6 & 22.8   & \cellcolor[HTML]{FFFFFF}29.2 \\ \hline
RAGS          & 19.8    & 2.6  & 17.8   & 29.6                         \\ \hline
\end{tabular}
\caption{Comparison of BAGS with Random BAGS (RAGS), where bin assignments are randomized rather than based on instance frequency.}
\label{table:rags}
\end{table}

\subsection{Extensions to BAGS}
We explored multiple extensions to the original BAGS framework, as detailed in Section~\ref{sec:bags_ext}, and their results are summarized in Table \ref{table:bags-mod}. Each modification targeted specific category sets. For instance, incorporating class weights improved the mAP for rare and common categories while slightly reducing performance for frequent categories.

Based on these findings, we propose a hybrid approach: employing softmax with class weights for rare and common categories, and standard softmax for frequent categories. This hybrid approach achieved an overall mAP of 24.5\%, surpassing the original BAGS method.

\begin{table}[!h]
\centering
\begin{tabular}{|l|l|l|l|l|}
\hline
Modification     & Overall                      & Rare                         & Common                       & Frequent                     \\ \hline
Original         & 24.0                         & 15.6                         & 22.8                         & \cellcolor[HTML]{9E9E9E}29.2 \\ \hline
5-Bins           & 24.2                         & 15.6                         & 23.1                         & 29.1                         \\ \hline
Clustered        & 24.0                         & 15.0                         & 23.0                         & 29.1                         \\ \hline
Focal Loss       & 24.0                         & \cellcolor[HTML]{9E9E9E}16.4 & 23.2                         & 28.2                         \\ \hline
Class Weighted   & \cellcolor[HTML]{9E9E9E}24.3 & 16.1                         & \cellcolor[HTML]{9E9E9E}23.6 & 28.7                         \\ \hline
\rowcolor[HTML]{EFEFEF}
Hybrid           & \cellcolor[HTML]{9E9E9E}24.5 & 16.6                         & 23.1                         & 29.4                         \\ \hline
\end{tabular}
\caption{mAP values on the LVIS-V1 validation set for various extensions to BAGS. Highlighted cells indicate the best results for each category set (excluding the hybrid approach).}
\label{table:bags-mod}
\end{table}

\begin{figure}
    \centering
    \includegraphics[width=\columnwidth]{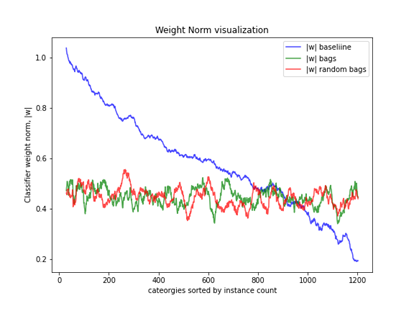}
    \caption{Classifier weights visualization.}
    \label{fig:my_label}
\end{figure}

\subsection{Metric Learning}

\subsubsection{Center Loss}

We trained the Faster R-CNN model with Center Loss for 12 epochs, using a feature dimension of 1024 for center calculation and KNN inference. Initially, the model was trained on a subset of 30 randomly selected categories from the LVIS-V1 dataset. We experimented with multiple values for the weight assigned to the center loss term ($\lambda$), specifically 0.1 and 0.01. As outlined in the previous section, both softmax-based inference and KNN inference were evaluated. The results for this experiment are provided in the first four rows of Table \ref{table:knn}.

For the 30-category subset, we observed that KNN inference improved the mAP for rare and common classes compared to softmax inference. However, overall mAP remained higher for the standard softmax inference approach.

We also trained the model on the entire dataset with $\lambda = 0.01$. The results for this setup are shown in the last row of Table \ref{table:knn}, and similar trends were observed.

\begin{figure}[ht]
    \centering
    \includegraphics[width=0.3\textwidth]{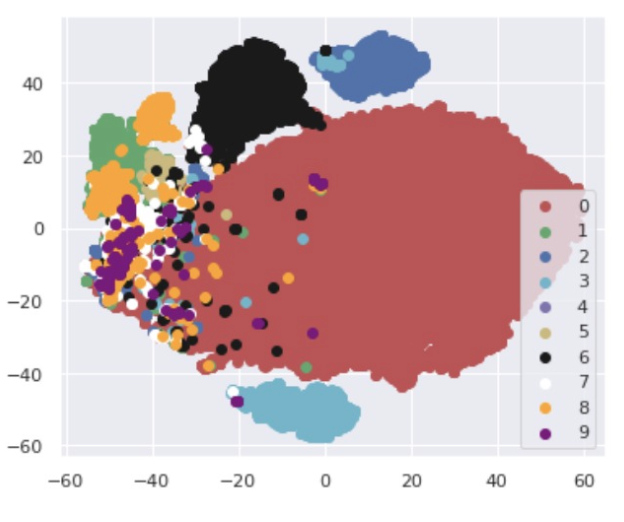}
    \caption{t-SNE visualization of category features for 9 categories, sorted in decreasing order of frequency. Tail class features are scattered and lack tight clusters}
    \label{fig:tsne}
\end{figure}

To better understand the observed performance, we visualized the class features in a 2D space using t-SNE transformations, as shown in Figure \ref{fig:tsne}.

\begin{figure}[ht]
    \centering
    \includegraphics[width=0.45\textwidth]{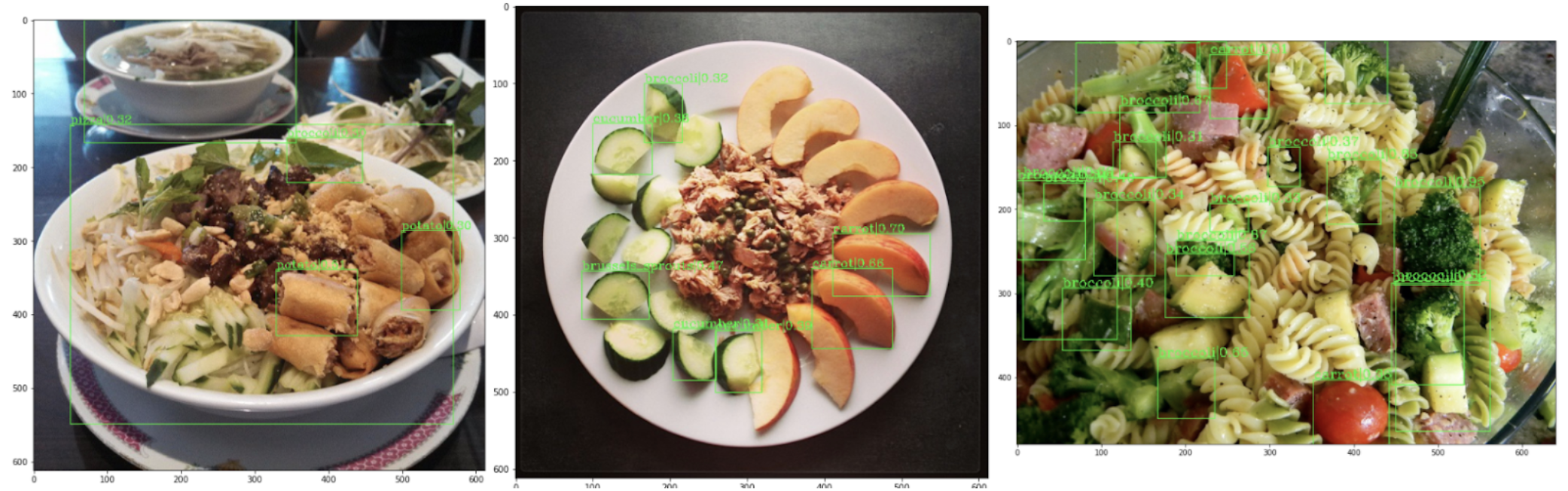}
    \caption{Variations in color, texture, and shape among instances of the "cucumber" class.}
    \label{fig:cucumber}
\end{figure}

Feature visualization revealed that tail class features did not cluster as tightly as expected, appearing scattered with high variance. This behavior is intuitive; certain tail classes, such as "cucumber" (illustrated in Figure \ref{fig:cucumber}), exhibit significant intra-class variations in shape, size, and texture across different instances. Additionally, the dataset contains 1203 categories, and inherent similarities between some categories further hinder the formation of completely isolated clusters. Consequently, nearest-neighbor classification is less effective due to the overlap of feature clusters, explaining the lower mAP values.

\begin{table}[ht]
\centering
\begin{tabular}{|l|l|l|l|l|l|}
\hline
Method & $\lambda$ & Overall & Rare                        & Common & Frequent \\ \hline
Softmax          & 0.01               & 14.7    & 0.0                         & 13.4   & 23.6     \\ \hline
Softmax          & 0.1                & 5.8     & 0.0                         & 1.7    & 13.5     \\ \hline
KNN              & 0.01               & 9.8     & \cellcolor[HTML]{B7B7B7}1.5 & 5.6    & 19.0     \\ \hline
KNN              & 0.1                & 2.4     & 0.0                         & 0.0    & 6.6      \\ \hline
\rowcolor[HTML]{EFEFEF} 
KNN  & 0.01         & 5.0     & 0.0                         & 0.01   & 12.6     \\ 
\rowcolor[HTML]{EFEFEF} 
(all data) &  & & & & \\ \hline
\end{tabular}
\caption{Comparison of inference of models trained with Center Loss with specified weight $\lambda$ and Cross Entropy Loss using KNN and Softmax methods. The first four rows correspond to models trained on a 30 category subset of data. The last row corresponds to the model trained on the entire dataset with $\lambda = 0.01$.}
\label{table:knn}
\end{table}

\section{Future Work}

Metric learning holds significant potential for addressing the long-tail object detection challenge, but further exploration is necessary to fully understand its implications. Specifically, there is a need for a theoretical investigation of the high-dimensional feature space. Nearest-neighbor inference may prove suboptimal due to the potential multi-form nature of the feature distribution, necessitating consideration of multiple centers. Recent advancements in multi-modal foundational models and use of Generative AI Approaches could be explored to address class imbalance challenge, though such exploration is beyond the scope of this study.

Another promising direction involves representing tail class features as linear or nonlinear combinations of head class features. This approach could leverage the abundant head class data to enhance the representation and results for tail classes.

Additionally, more aggressive data augmentation strategies could be employed. For example, segmentation masks of objects could be used to replicate instances across diverse backgrounds and poses, effectively increasing the instance frequency of tail classes.

Lastly, we propose leveraging segmentation masks available in the LVIS dataset to enhance model training. Utilizing these masks could improve both bounding box localization and classification accuracy, leading to more robust performance across all category types.

\bibliography{references}

% \bibliography{references}

\end{document}